\documentclass[tablecaption=bottom,wcp]{jmlr} % W&CP article

\usepackage{booktabs}
\usepackage[load-configurations=version-1]{siunitx} % newer version
\theorembodyfont{\upshape}
\theoremheaderfont{\scshape}
\theorempostheader{:}
\theoremsep{\newline}

\jmlrproceedings{AABI 2018}{1st Symposium on Advances in Approximate Bayesian Inference, 2018}

\title[Disentangled Dynamic Representations from Unordered~Data]{Disentangled Dynamic Representations from Unordered~Data}

\usepackage{subcaption}

\author{\Name{Leonhard Helminger} \Email{leonhard.helminger@inf.ethz.ch}
	\addr ETH Zurich
	\AND
	\Name{Abdelaziz Djelouah} \Email{aziz.djelouah@disneyresearch.com}  
	\addr Disney Research
	\AND
	\Name{Markus Gross}
	\Email{grossm@inf.ethz.ch } 
	\addr ETH Zurich
	\AND
	\Name{Romann M.\ Weber}
	\Email{romann.weber@disneyresearch.com}
	\addr Disney Research
}

\begin{document}
	
	\maketitle
	
	\begin{abstract}
		We present a deep generative model that learns disentangled static and dynamic representations of data from unordered input. Our approach exploits regularities in sequential data that exist regardless of the order in which the data is viewed.  The result of our factorized graphical model is a well-organized and coherent latent space for data dynamics.  We demonstrate our method on several synthetic dynamic datasets and real video data featuring various facial expressions and head poses.
	\end{abstract}
	
	%\begin{keywords}
	%	Generative Model, Disentangled Representations, VAE
	%\end{keywords}

	\section{Introduction}
	\label{sec:intro}
	Unsupervised learning of disentangled representations is
	gaining interest as a new paradigm for data analysis.
	In the context of video, this is usually
	framed as learning two separate representations:
	one that varies with time and one that does not. 
	In this work we propose a deep generative model to learn
	this type of disentangled representation with an approximate variational posterior
	factorized into two parts to capture both static and dynamic 
	information. 
	Contrary to existing methods that mostly rely on recurrent 
	architectures, our model uses only random pairwise comparisons of observations 
	to infer information common across the data.
	Our model also includes a flexible prior that learns a distribution
	of the dynamic part given the static features. 
	As a result, our model can sample this low-dimensional 
	latent space to synthesize new unseen combinations
	of frames.

	\section{The Model}
	%Let $\mathbf{x}_{1:T} = (\mathbf{x}_1, \dots, \mathbf{x}_T)$ be a sequence of length $T$ and $p\left(\mathbf{x}_{1:T}\right)$ the corresponding probability distribution. In this work, \textcolor{red}{cross-ref to mandt} we model the data as latent variable model, that separates the static information (e.g. the face of a person) from the dynamic information (e.g. the expression). 
	
\begin{figure}[!htb]
		\hspace{-1.25cm}
		\vspace{-0.4em}
		\begin{minipage}{.5\columnwidth}
			\centering
			\includegraphics[scale=0.42]{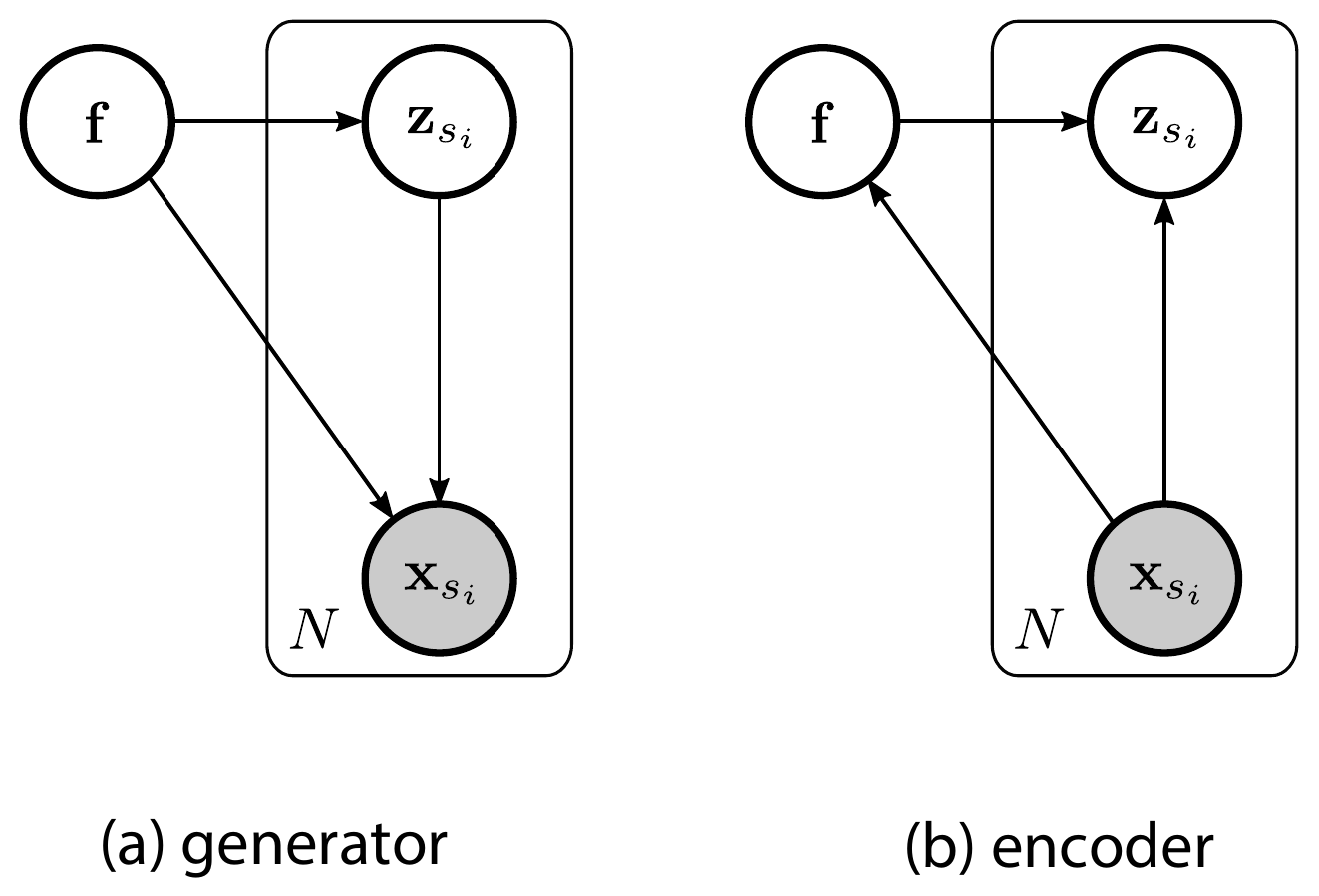}
			\caption{\small Graphical~Models}
			\label{fig:graph_model}
		\end{minipage}
		\hspace{-.75cm}
		\begin{minipage}{.6\textwidth}
			\centering
			\includegraphics[width=1.\linewidth]{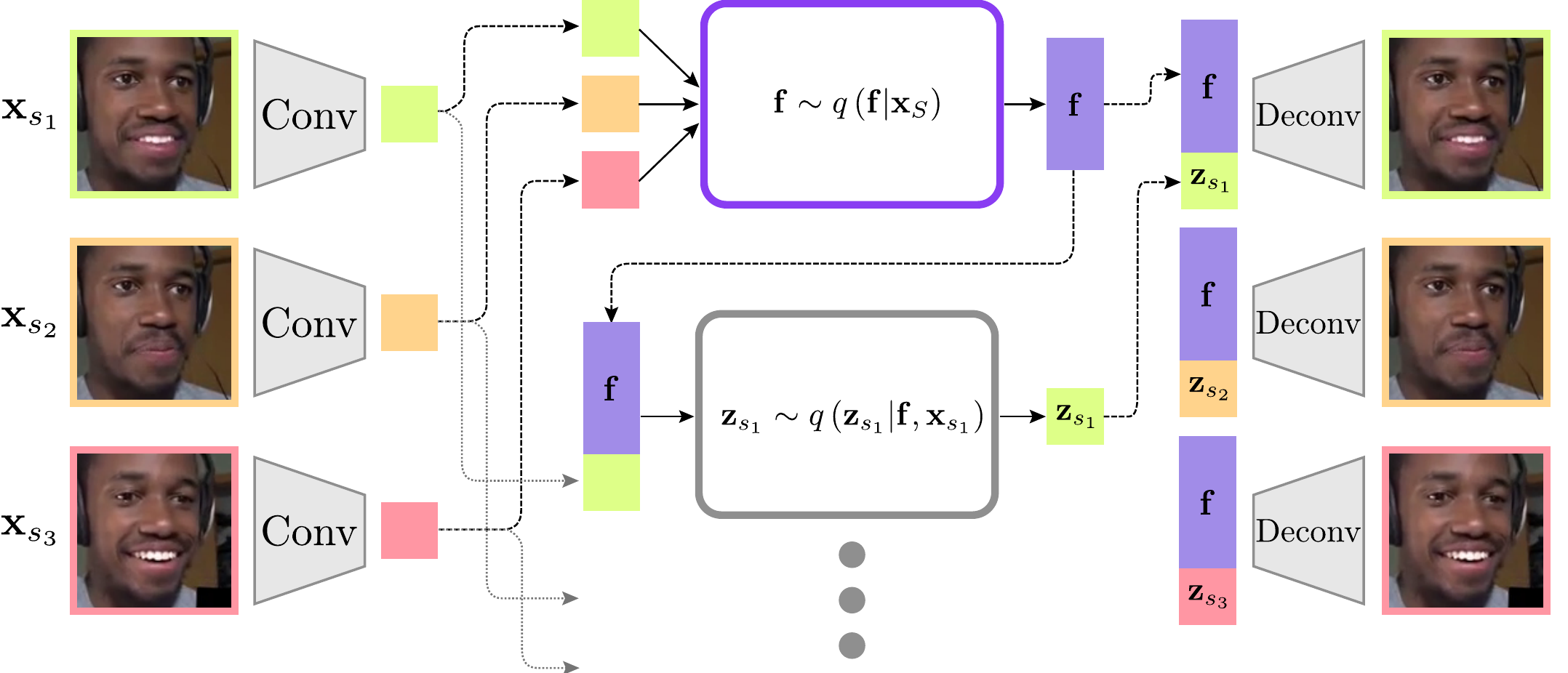}
			\caption{\small Model visualization}
			\label{fig:our_model}
		\end{minipage}
	\end{figure}

	Let $\mathbf{x}_{1:T} = (\mathbf{x}_1, \dots, \mathbf{x}_T)$ be a data sequence of length $T$ and $p\left(\mathbf{x}_{1:T}\right)$ its corresponding probability distribution. 
	We assume that each sequence $\mathbf{x}_{1:T}$ is generated from a random process involving latent variables $\mathbf{f}$ and $\mathbf{z}_{1:T}$. The generation process, as illustrated in Figure (\ref{fig:graph_model}a), can be explained as follows: \textit{(i)} a vector $\mathbf{f}$ is drawn from the prior distribution $p_\theta\left(\mathbf{f}\right)$, \textit{(ii)} $T$ i.i.d.\ latent variables $\mathbf{z}_{1:T}$ are drawn from the sequence-dependent but time-independent conditional distribution $p_\theta\left(\mathbf{z}\:|\:\mathbf{f}\right)$, \textit{(iii)} $T$ i.i.d.\ observed variables $\mathbf{x}_{1:T}$ are drawn from the conditional distribution $p_\theta\left(\mathbf{x}\:|\:\mathbf{z}, \mathbf{f}\right)$.\\
	%---------------------
	
	%Further, we drop the time dependencies between the elements of the sequence, and assume that each element is independent of its predecessors. 
	
	%\begin{figure}[htbp]
	%\floatconts
	%{fig:subfigex}
	%{\caption{A visualization of the encoder and generator as %graphical models.}}
	%{%
	%		\subfigure[generator]{\label{fig:gm_generator}%
	%			\includegraphics[width=0.2\linewidth]{images/genera%tor_new.pdf}}%%
	%\hspace{4em}
	%\subfigure[encoder]{\label{fig:gm_encoder}%
	%	\includegraphics[width=0.2\linewidth]{images/encoder_new.pdf}}
	%}
	%\end{figure}

	\textbf{Generative Model:} The generative model that describes the generation process above is given by 
	%\vspace{-0.25cm}
	\begin{equation*}
	p_\theta\left(\mathbf{x}_{1:T}, \mathbf{f}, \mathbf{z}_{1:T}\right) = p_\theta\left(\mathbf{f}\right)\prod_{t=1}^T p_\theta\left(\mathbf{x}_t\:|\: \mathbf{f}, \mathbf{z}_t\right) p_\theta\left(\mathbf{z}_t \:|\:\mathbf{f}\right),
	\end{equation*}
	%\vspace{-0.25cm}
	where $\mathbf{f}$ and $\mathbf{z}_t$ are the latent variables that contain the static and dynamic information of each element, respectively. The parameters of the generative model are denoted as  $\mathbf{\theta},$ and the RHS terms are formulated as follows:
	\begin{align*}
	p_\theta\left(\mathbf{f}\right) &= \mathcal{N}\left(\mathbf{f}\:|\:\mathbf{0}, \mathbf{1}\right),\\
	p_\theta\left(\mathbf{z}\:|\:\mathbf{f}\right) &= \mathcal{N}\left(\mathbf{z}\:|\: h_{\mu_z}\left(\mathbf{f}\right), \mbox{diag}\left(h_{\sigma^2_z}\left(\mathbf{f}\right)\right)\right)  \\
	p_\theta\left(\mathbf{x}\:|\:\mathbf{f}, \mathbf{z}\right) &= \operatorname{Ber}_p\left(\mathbf{x}\:|\:h_x\left(\mathbf{f}, \mathbf{z}\right)\right),
	\end{align*}
	where $p_\theta\left(\mathbf{f}\right)$ is a standard normal distribution and $p_\theta\left(\mathbf{z}\:|\:\mathbf{f}\right)$ is a multivariate normal distribution,
	parameterized by two neural networks $h_{\mu_z}$ and $h_{\sigma^2_z}.$ 
	The likelihood $p_\theta\left(\mathbf{x}\:|\:\mathbf{f}, \mathbf{z}\right)$ is a Bernoulli distribution parameterized by a neural network $h_x$. Experimentally, this leads to
	sharper results than using a Normal distribution. \\%with inputs $\mathbf{f}$ and $\mathbf{z}$.\\

	%\textit{Consider the case where $\mathbf{x}_{1:T}$ is a video sequence and $\mathbf{x}_t$ is a corresponding frame. The generative model is factorized in such a way, such that the static information e.g. the general appearance of the face is encoded in the latent vector $\mathbf{f}$ and the dynamics e.g. the expression is encoded in the latent variable $\mathbf{z}_t$. Since we are interested in the general separation of the fixed and the dynamic part of a frame, we intentionally remove the time dependencies and treat the frames as independent given the identity $\mathbf{f}$.}\\
	
	\textbf{Inference Model:}
	To overcome the problem of intractable inference with the true posterior, we define an approximate inference model, $q_\phi\left(\mathbf{f}, \mathbf{z}_{1:T}\:|\: \mathbf{x}_{1:T}\right).$  We train the generative model within the VAE framework proposed by \cite{DBLP:journals/corr/KingmaW13}.
	
	To successfully separate the static from the dynamic information, 
	the model needs to know which information is common among $\mathbf{x}_{1:T}$. 
	While a sequence could be arbitrary long, we randomly sample $N$ frames, $\mathbf{x}_S = \left(x_{s_1}, \dots, x_{s_N}\right),$ from the sequence, whose pairwise comparison helps us compute the encoding for the static information $\mathbf{f}$.\\

	We now consider the factorized inference model as depicted in Figure (\ref{fig:graph_model}b):
	%\vspace{-0.25cm}
	\begin{align*}
	q_\phi\left(\mathbf{f}, \mathbf{z}_{s_{1:N}} | \mathbf{x}_S\right) &= q_\phi\left(\mathbf{f}\:|\: \mathbf{x}_S\right)\prod_{i=1}^N q_\phi\left(\mathbf{z}_{s_i}\:|\:\mathbf{f}, \mathbf{x}_{s_i}\right) \\
	q_\phi\left(\mathbf{f} \:|\: \mathbf{x}_S\right) &= \mathcal{N}\left(\mathbf{f}\:|\: g_{\mu_f}\left(\mathbf{x}_S\right), \mbox{diag}\left(g_{\sigma^2_f}\left(\mathbf{x}_S\right)\right)\right) \\
	q_\phi\left(\mathbf{z} \:|\: \mathbf{f}, \mathbf{x}\right) &= \mathcal{N}\left(\mathbf{z}\:|\: g_{\mu_z}\left(\mathbf{f}, \mathbf{x}\right), \mbox{diag}\left(g_{\sigma^2_z}\left(\mathbf{f}, \mathbf{x}\right)\right)\right),
	\end{align*}
	where the posteriors over $\mathbf{f}$ and $\mathbf{z}$ are multivariate normal distributions parameterized by neural networks $g_{\mu_z}\left(\mathbf{f}, \mathbf{x}\right)$ and $g_{\sigma^2_z}\left(\mathbf{f}, \mathbf{x}\right)$. The inference model parameters are denoted by $\phi.$

	%\textit{We further pairwise concatenate the frames in the encoder, to remove the temporal order between the elements.
	%This simplification gives us the following two advantages:
	%Firstly, we are able to utilize convolution layers to extract common parts of the frames. This leads to a clean and well structured encoder.
	%Second, the sampling step allows us to increase the size of the dataset exponentially. 
	%Consider the sequence $\mathbf{x}_{1:T}$ of length $T$. 
	%By randomly sampling from $\mathbf{x}_{1:T}$, we can generate a new dataset $\mathcal{D}$ of size $|\mathcal{D}| = 3^T$.}\\
	
	In the inference model $q\left(\mathbf{f}\: | \: \mathbf{x}_S\right)$ we learn the static information of $\mathbf{x}_S$. 
	We achieve this by using the same convolutional layer for every concatenated pair of frames $\left(\mathbf{x}_{s_j}, \mathbf{x}_{s_i}\right) \in \mathbf{x}_S.$
	Through this architecture, the encoder learns only the common information of frames $\mathbf{x}_S$.\footnote{For more details see Appendix Figure (\ref{fig:encoder_f}).}  %The second term of the inference model is the conditional distribution $q\left(\mathbf{z}\:|\: \mathbf{f}, \mathbf{x}\right)$.
	%Consider the case where $\mathbf{x}_{1:T}$ is a video sequence and $\mathbf{x}_S$ is a set of randomly sampled frames from a single sequence. 
	%The conditional distribution $q\left(\mathbf{f}\: | \: \mathbf{x}_S\right)$ learns to encode the static information of the frames  $\mathbf{x}_S$. 
	%We achieve this, by using the same convolutional layer for every concatenated pair of frames $\left(\mathbf{x}_{s_j}, \mathbf{x}_{s_i}\right)$. 
	%Through this architecture, the encoder is not able to learn specific details of a frame, but rather the common information of frames $\mathbf{x}_S$. 
	
	%\begin{figure}
	%	\centering
	%	\includegraphics[width=0.7\linewidth]{images/test_model_prob.pdf}
	%	\caption{Visualization of our model}
	%	\label{fig:model}
	%\end{figure}
	
	\textbf{Learning:} The variational lower bound for our model is given by 
	%\vspace{-0.25mm}
	\begin{align*}
	\mathcal{L} &= \mathbb{E}_{q_\phi\left(\mathbf{f}\:|\:\mathbf{x}_S\right)}\left[
	\sum_{\mathbf{x} \in \mathbf{x}_S} \mathbb{E}_{q_\phi\left(\mathbf{z}\:|\: \mathbf{f}, \mathbf{x}\right)} \left[
	\operatorname{log}p_\theta\left(\mathbf{x}\:|\:\mathbf{f}, \mathbf{z}\right) -
	\operatorname{D}_{KL}\left(q_\phi\left(\mathbf{z}\:|\:\mathbf{f}, \mathbf{x}\right) \: || \: p_\theta\left(\mathbf{z}\:|\: \mathbf{f}\right)\right)
	\right]
	\right] \\
	& \hspace{19em} - \operatorname{D}_{KL}\left(q_\phi\left(\mathbf{f}\:|\: \mathbf{x}_S\right)\:||\:p_\theta\left(\mathbf{f}\right)\right),
	\end{align*}
	which we optimize with respect to the variational parameters $\theta$ and $\phi.$
	\section{Related Work}

	%	Unsupervised learning of disentangled representation
	%	can be related to learning structured latent space
	%	using variational auto-encoder. 
	%	For example~\cite{DBLP:conf/nips/0002DWAD16} propose 
	%	different architectures of generative models to learn a 
	%	richer distribution in the latent space. 
	%	Using multiple stochastic layers,
	%	\cite{DBLP:conf/nips/SonderbyRMSW16} also 
	%	learn a structured latent space with a deep 
	%	generative model.
	%	More recently~\cite{DBLP:journals/corr/TomczakW17} 
	%	explore the learning of a flexible prior, 
	%	by a mixture of variational posteriors conditioned 
	%	on learnable pseudo-data.
	
	Unsupervised learning of disentangled representations
	can be related to modeling context or hierarchical structure in datasets. 
	In particular, our approach invites comparison to the ``neural statistician'' of \cite{DBLP:journals/corr/EdwardsS16}, whose context variable closely corresponds to our static encoding, although our model has a different dependence structure.
	%\cite{DBLP:journals/corr/abs-1807-08919} 
	%subsample from different datasets, 
	%and apply the model on factorial and hierarchical structured 
	%datasets. 
	%	Here a structured PixelCNN 
	%	(\cite{DBLP:journals/corr/OordKVEGK16})
	%	prior for the context latent variable is used.
	%	{\color{red}argument why we are different?}
	
	On sequential data, 
	\cite{DBLP:conf/nips/HsuZG17} propose a factorized 
	hierarchical variational auto-encoder using a lookup table 
	for different means, while 
	\cite{DBLP:conf/icml/LiM18a} 
	condition a component of the factorized prior on the full ordered sequence. 
	%The major difference of our approach is the latent 
	%variables that capture the dynamics depend on each other. 
	%In contrast, we force the model to remove the temporal 
	%dependencies between the frames.
	\cite{DBLP:conf/nips/DentonB17} use an adversarial loss to 
	factor the latent representation of a video frame in a 
	stationary and temporally varying component. 
	\cite{DBLP:journals/corr/TulyakovLYK17} 
	introduce a GAN that produces video clips 
	by sequentially decoding a sample vector that consists of two parts: a sample from the motion subspace and a sample from a content subspace. 
	%This is different from our work as we explore
	%the usage of variational autoencoders instead
	%of GANs.
	In video generation, other directions can also be explored 
	by decomposing the learned representation into deterministic 
	and stochastic
	(\cite{DBLP:journals/corr/abs-1802-07687}).

	\begin{figure}[t]
		\floatconts
		%\label{fig:results}
		{fig:result1}
		{\vspace{-0.4cm}\caption{Visualizations of learned dynamic data manifold of the two-dimensional latent space $\mathbf{z}$ for the MMNIST and Sprite dataset.}}
		{	\vspace{-0.5cm}
			\subfigure[MMNIST]{\label{fig:ls_mmnist}
				\includegraphics[width=0.25\linewidth]{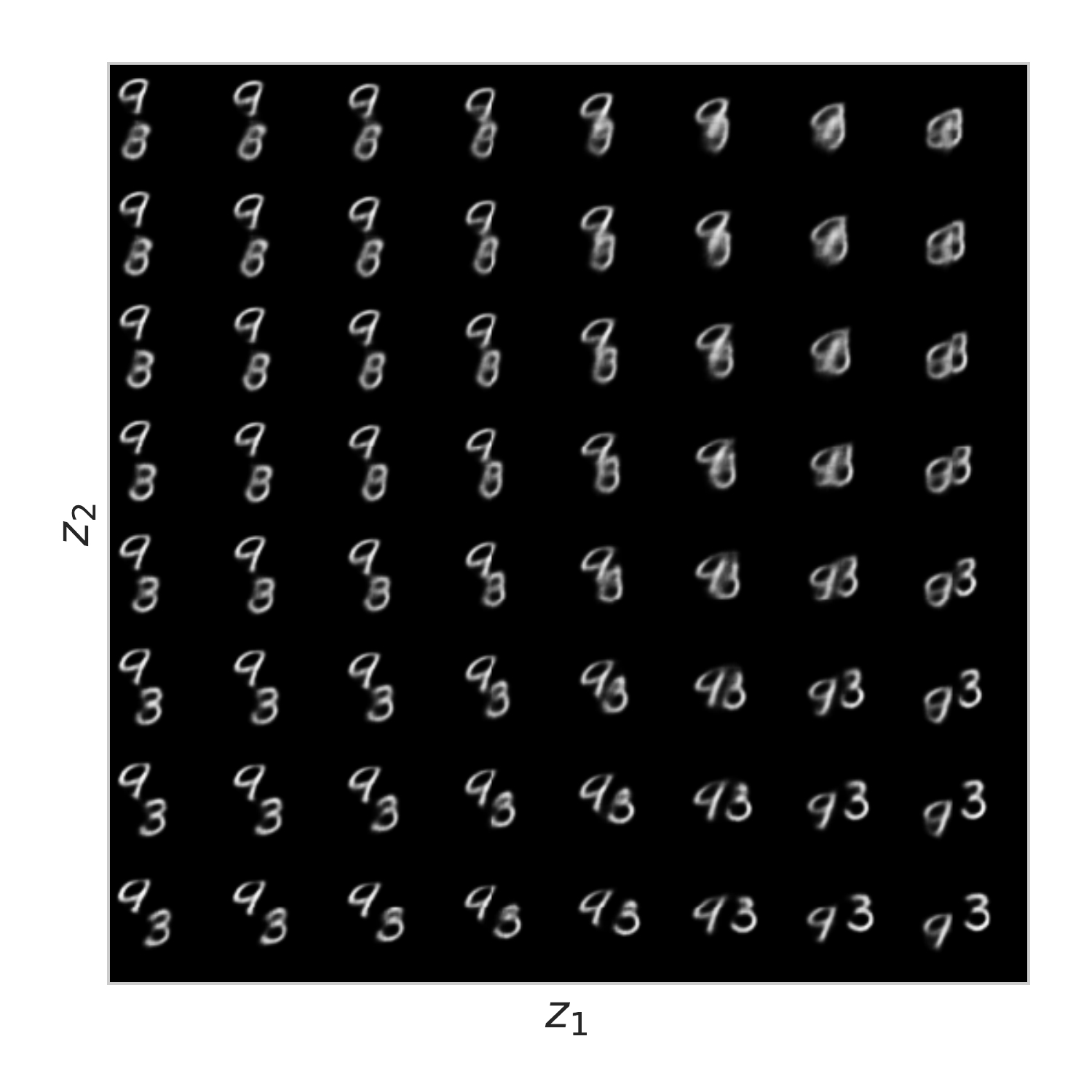}%
				\hspace{-1.0em}
				\includegraphics[width=0.25\linewidth]{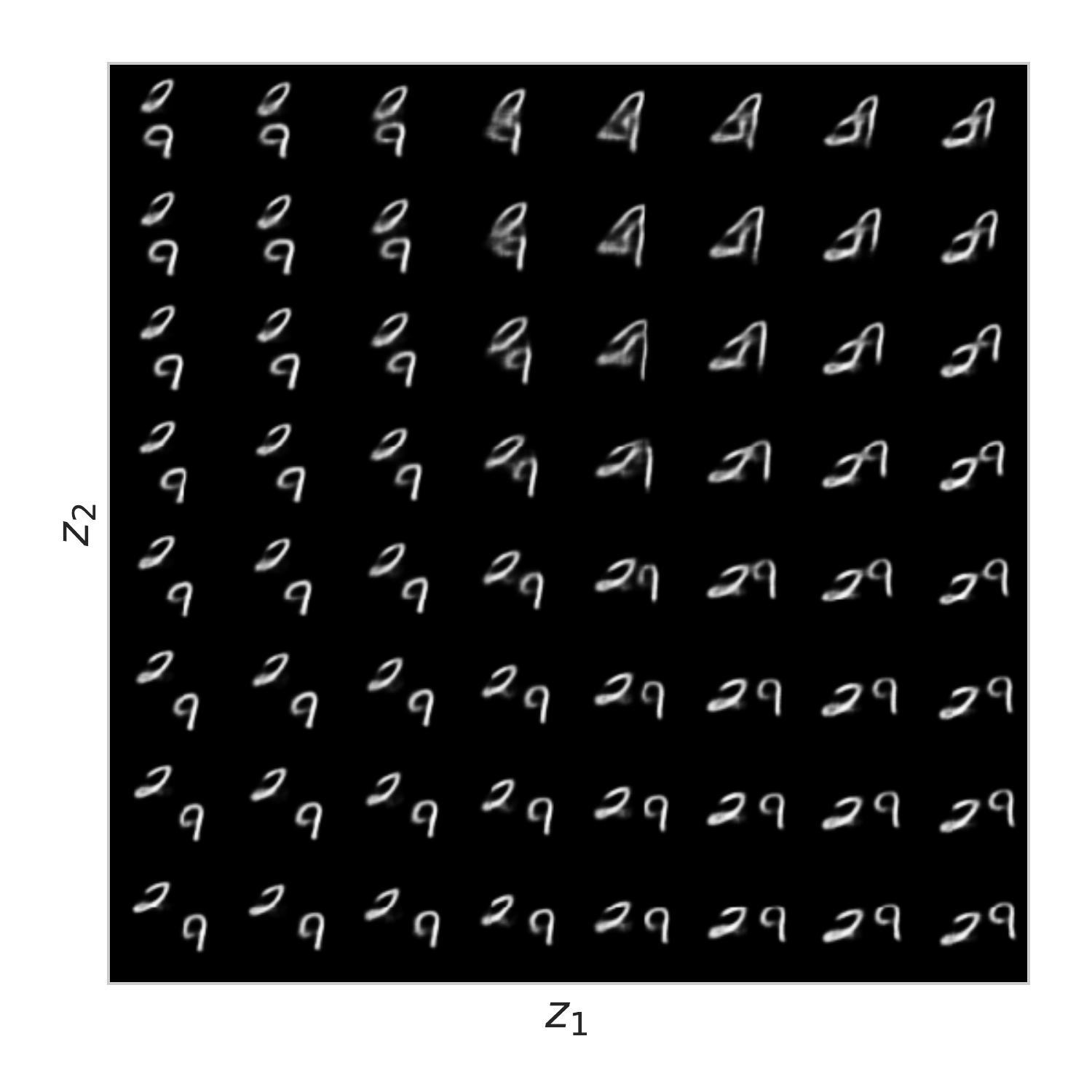}
				\vspace{-5.5cm}
			}
			\hspace{-0.6em}
			\subfigure[Sprite]{\label{fig:ls_sprite}
				\includegraphics[width=0.25\linewidth]{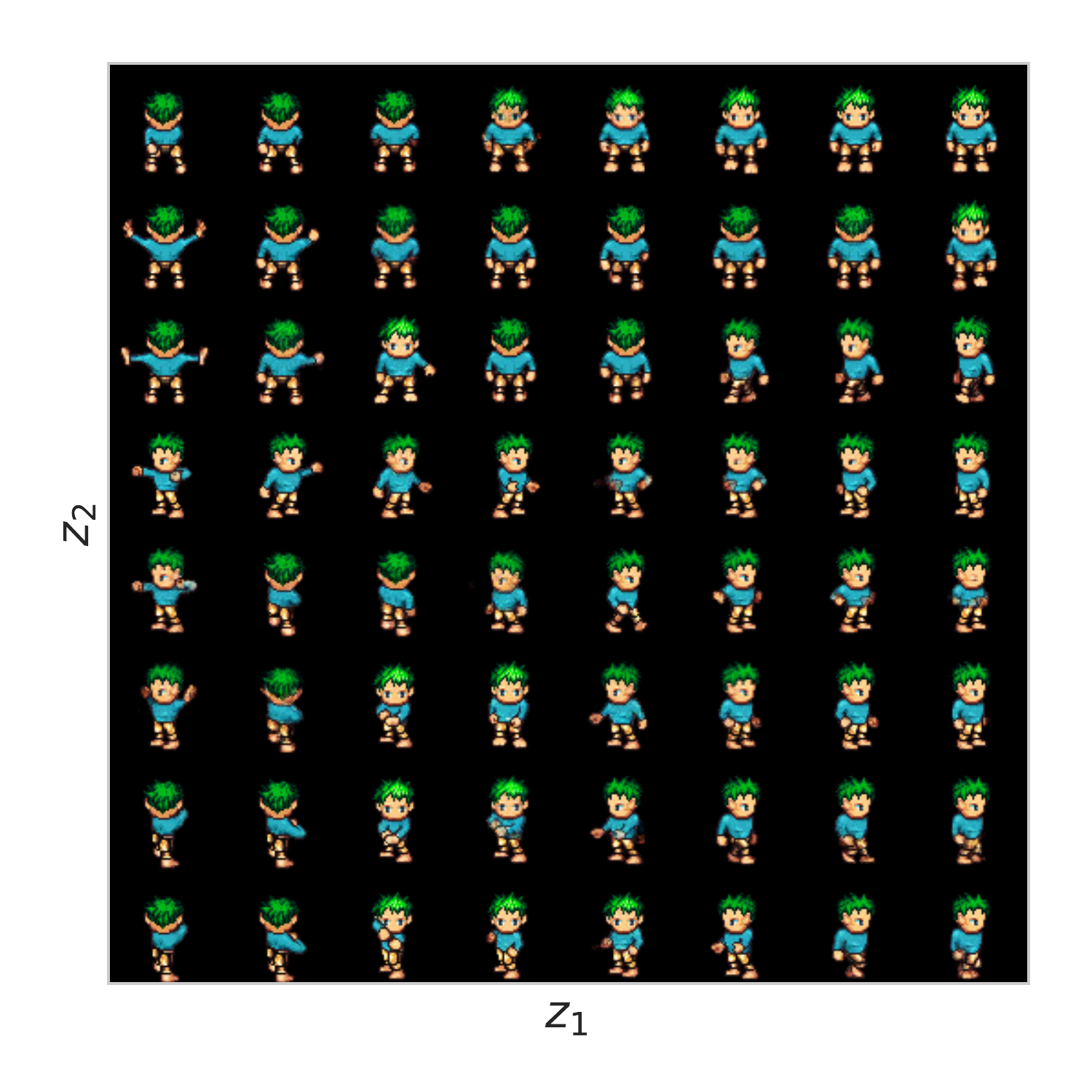}%
				\hspace{-1.0em}
				\includegraphics[width=0.25\linewidth]{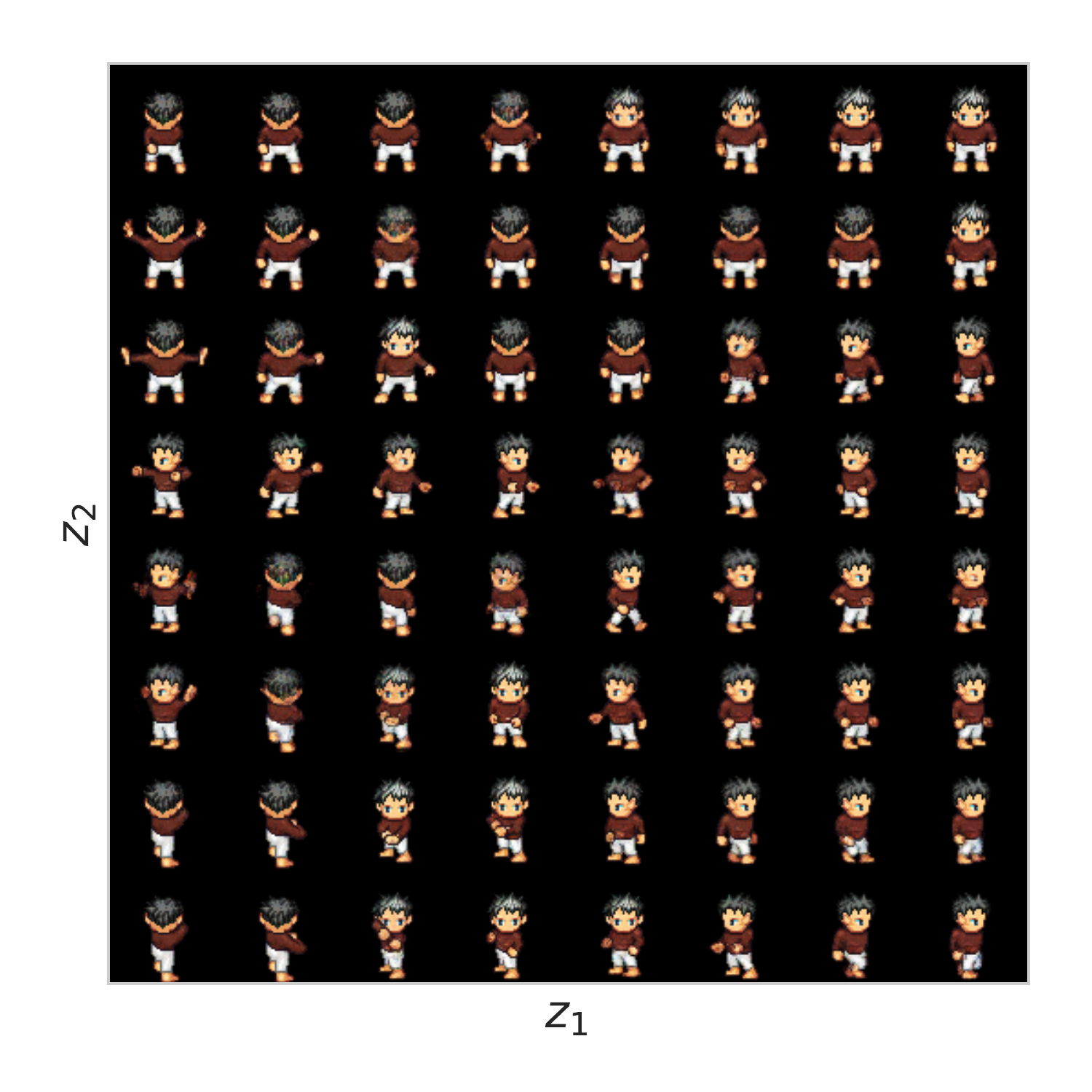}
				\vspace{-5.5cm}
			}
		}
	\end{figure}
	
	\section{Experiments}
	We evaluate our model on two synthetic datasets 
	Sprites (\cite{DBLP:conf/icml/LiM18a}) and Moving MNIST (\cite{DBLP:conf/icml/SrivastavaMS15}),
	and a real one, Aff-Wild dataset (\cite{DBLP:journals/corr/abs-1804-10938}).
	The detailed description of the preprocessing on these
	datasets is provided in appendix.
	
	\begin{figure}[h]
		\floatconts
		{fig:result2}
		{\vspace{-0.4cm}\caption{(a) Visualizations of learned dynamic data manifold of the two-dimensional latent space $\mathbf{z}$ for the AFF-Wild dataset (b) Plot of the first dimension of the encoded dynamics of video frames for a single sequence with some of the corresponding frames. }}
		{%
			\subfigure[AFF-Wild Dataset]{\label{fig:ls_aff}
				\includegraphics[width=0.26\linewidth]{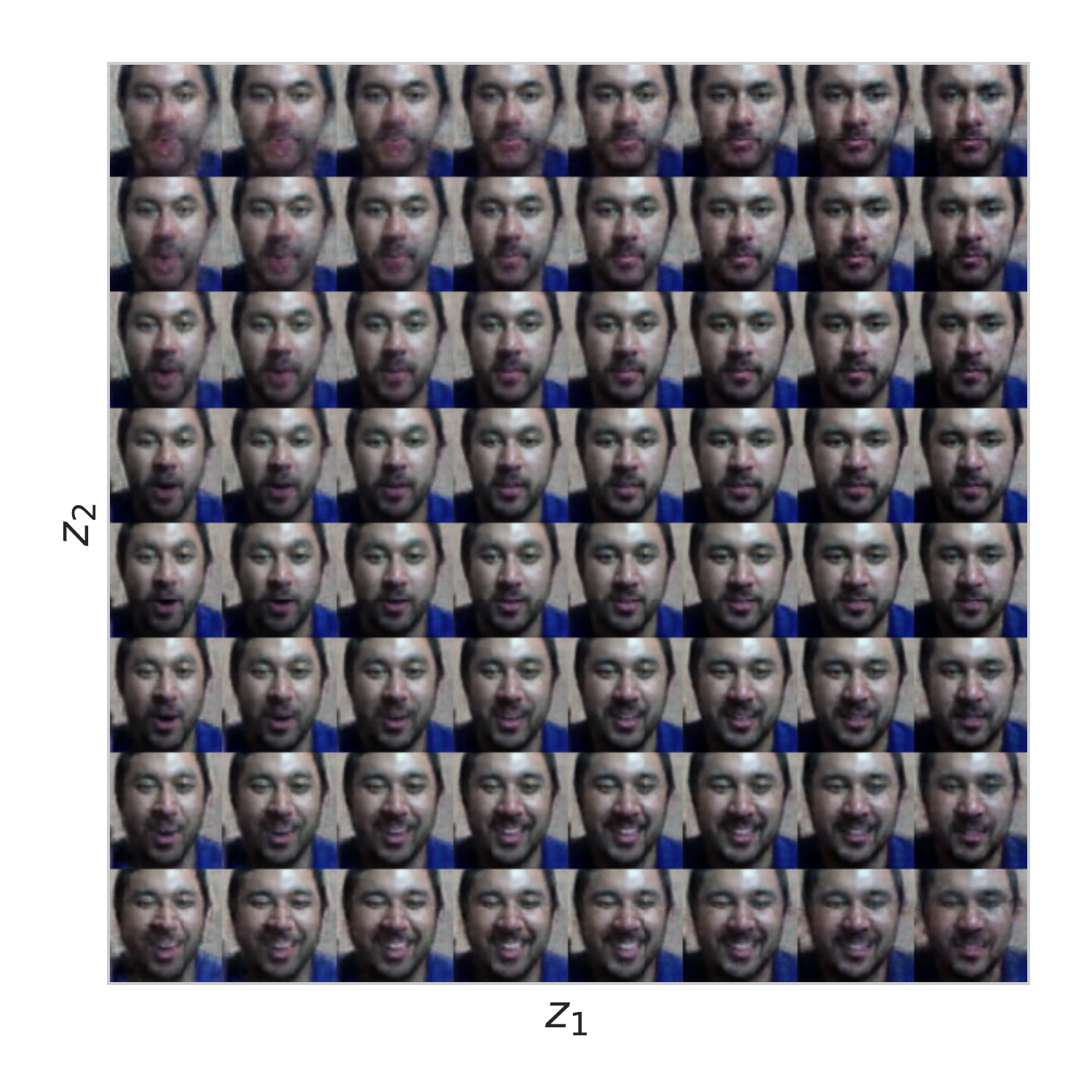}
				\hspace{-1.0em}
				\includegraphics[width=0.26\linewidth]{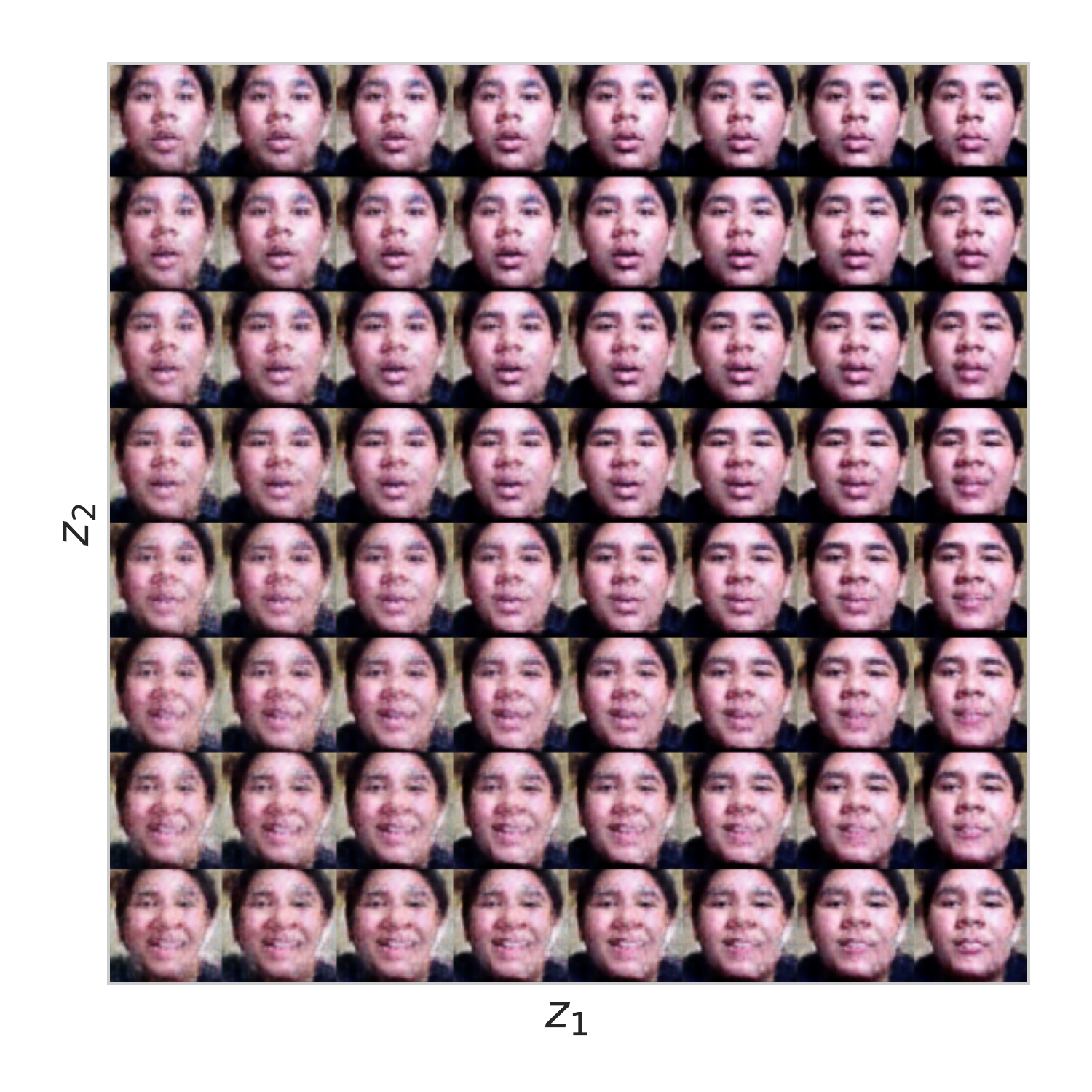}%
			}
			\hspace{-0.5em}
			\subfigure[Encoded Video Sequence]{\label{fig:timeline}
				\includegraphics[width=0.45\linewidth]{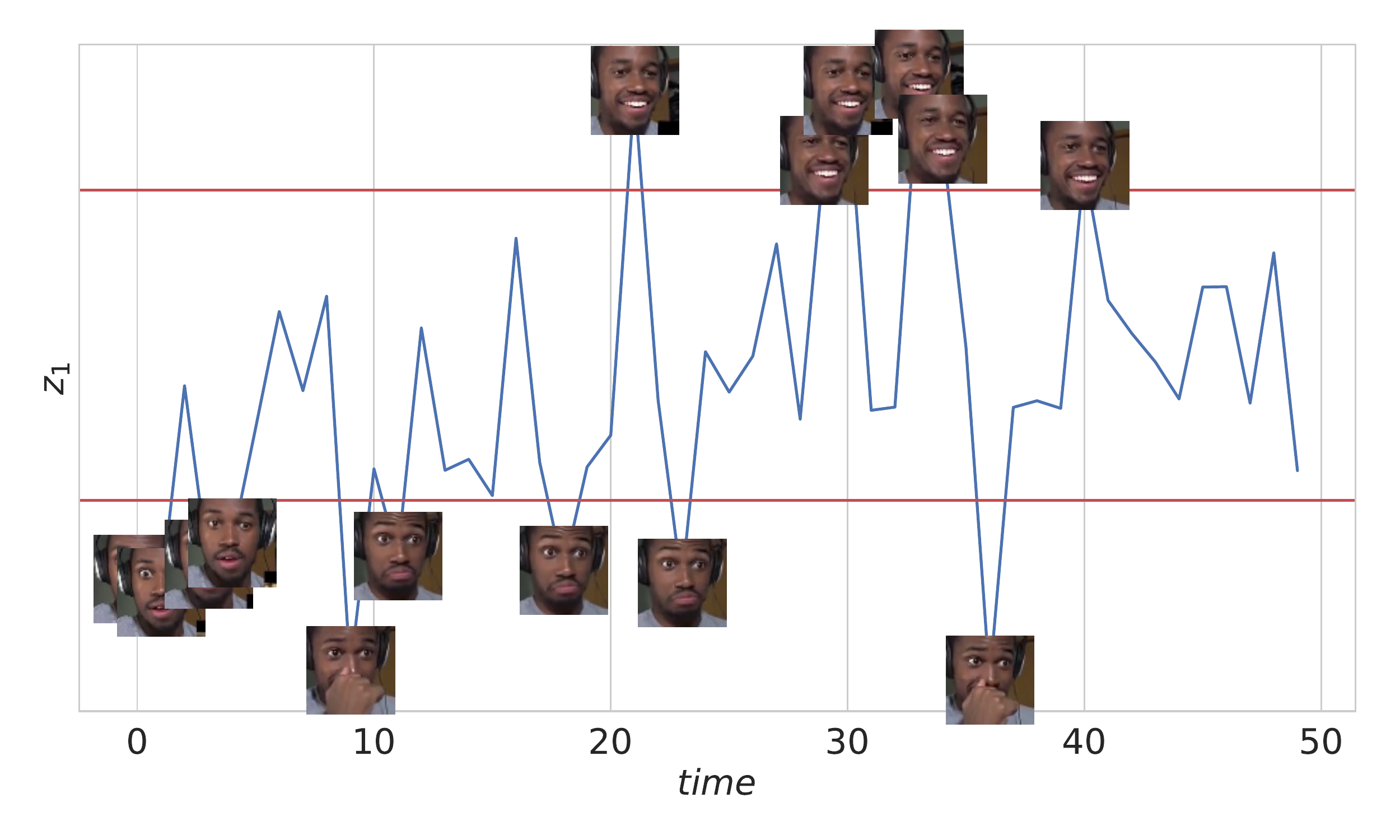}%
			}
		}
	\end{figure}
	
	\textbf{Qualitative evaluation} 
	For all models in this section we use $N = 3$ frames and a 
	batch size of 120. We set the dimension of the latent space 
	$\mathbf{z}$ for the dynamic information to $2$. 
	We used ADAM with learning rate $1e^{-4}$ to optimize the 
	variational lower bound $\mathcal{L}$.
	To show the learned dynamics of the sequences of the 
	datasets, we fix $\mathbf{f}$ and visualize the decoded 
	samples from the grid in the latent space $\mathbf{z}$ 
	(Fig.~\ref{fig:result1}).
	In the case of the MMNIST, the digits and style 
	of the handwritten numbers are consistent over 
	the spanned space. The encoded dynamic 
	can be interpreted as the position and 
	orientation of the digits. 
	Similar observation holds for the sprites dataset, 
	but this time $\mathbf{z}$ encode the pose of the character. 
	In both cases we can note the coherence of 
	the dynamic space between different identities. 
	%	Especially interesting in this result is the fact, that for two different appearances of the character, the poses are located in the same area (relative to the mean) of the dynamic space. A reason for this behavior could be the coherent body form over all possible identities. \\
	
	\textbf{Application}: Even with just two dimensions,
	the latent space $\mathbf{z}$ captures 
	the dynamics of faces well, suggesting that it can be used for representing and analyzing 
	expressions in an unsupervised way. 
	To illustrate this concept, we plot one of the dynamic components
	of a sequence (Figure~\ref{fig:timeline}). 
	A naive analysis of this dynamic plot can already 
	extract some meaningful facial expressions from a specific 
	person. In this specific example, expressions we would call smiling and
	astonished.
	%    that 
	%    For a potential application of this 
	%    learned disentangled representation of dynamics, we consider 
	%    the result visualized in Figure (\ref{fig:ls_aff}). Even with 
	%    just two dimensions, the model is able to encode various emotions (e.g. smiling, astonished or neutral) in the latent space . These encodings of facial expressions could be used to classify the emotion of a person, in an unsupervised way. 
	
	\section{Discussion}
	In this work, we introduced a deep generative model that effectively learns disentangled static and dynamic representations from data without temporal ordering.  
	The ability to learn from unordered data is important
	as one can take advantage of the combinatorics 
	of randomly choosing pairwise 
	comparisons, to train models on small datasets.
	%An interesting consequence of using unordered data is that the combinatorics of randomly choosing pairwise comparisons makes it so that the model is unlikely to ever see the same set of examples twice over training
	%As shown in the experiments, the model learns, even without temporal coherency, a smooth and mean-ingfullatent space. 
	While in the current model the same frames are used to compute both
	the dynamic and static encodings, an interesting subject for future work would be to see if defining a distinct set of frames for the dynamic part would lead to a better separation.
	
	% {\color{red} Add some sort of concluding remarks here.}
	
	\newpage
	\bibliography{jmlr-sample}

\begin{thebibliography}{9}
\providecommand{\natexlab}[1]{#1}
\providecommand{\url}[1]{\texttt{#1}}
\expandafter\ifx\csname urlstyle\endcsname\relax
  \providecommand{\doi}[1]{doi: #1}\else
  \providecommand{\doi}{doi: \begingroup \urlstyle{rm}\Url}\fi

\bibitem[Denton and Fergus(2018)]{DBLP:journals/corr/abs-1802-07687}
Emily Denton and Rob Fergus.
\newblock Stochastic video generation with a learned prior.
\newblock \emph{CoRR}, abs/1802.07687, 2018.

\bibitem[Denton and Birodkar(2017)]{DBLP:conf/nips/DentonB17}
Emily~L. Denton and Vighnesh Birodkar.
\newblock Unsupervised learning of disentangled representations from video.
\newblock In \emph{{NIPS}}, pages 4417--4426, 2017.

\bibitem[Edwards and Storkey(2016)]{DBLP:journals/corr/EdwardsS16}
Harrison Edwards and Amos~J. Storkey.
\newblock Towards a neural statistician.
\newblock \emph{CoRR}, abs/1606.02185, 2016.

\bibitem[Hsu et~al.(2017)Hsu, Zhang, and Glass]{DBLP:conf/nips/HsuZG17}
Wei{-}Ning Hsu, Yu~Zhang, and James~R. Glass.
\newblock Unsupervised learning of disentangled and interpretable
  representations from sequential data.
\newblock In \emph{{NIPS}}, pages 1876--1887, 2017.

\bibitem[Kingma and Welling(2013)]{DBLP:journals/corr/KingmaW13}
Diederik~P. Kingma and Max Welling.
\newblock Auto-encoding variational bayes.
\newblock \emph{CoRR}, abs/1312.6114, 2013.

\bibitem[Kollias et~al.(2018)Kollias, Tzirakis, Nicolaou, Papaioannou, Zhao,
  Schuller, Kotsia, and Zafeiriou]{DBLP:journals/corr/abs-1804-10938}
Dimitrios Kollias, Panagiotis Tzirakis, Mihalis~A. Nicolaou, Athanasios
  Papaioannou, Guoying Zhao, Bj{\"{o}}rn~W. Schuller, Irene Kotsia, and
  Stefanos Zafeiriou.
\newblock Deep affect prediction in-the-wild: Aff-wild database and challenge,
  deep architectures, and beyond.
\newblock \emph{CoRR}, abs/1804.10938, 2018.

\bibitem[Li and Mandt(2018)]{DBLP:conf/icml/LiM18a}
Yingzhen Li and Stephan Mandt.
\newblock Disentangled sequential autoencoder.
\newblock In \emph{{ICML}}, volume~80 of \emph{{JMLR} Workshop and Conference
  Proceedings}, pages 5656--5665. JMLR.org, 2018.

\bibitem[Srivastava et~al.(2015)Srivastava, Mansimov, and
  Salakhutdinov]{DBLP:conf/icml/SrivastavaMS15}
Nitish Srivastava, Elman Mansimov, and Ruslan Salakhutdinov.
\newblock Unsupervised learning of video representations using lstms.
\newblock In \emph{{ICML}}, volume~37 of \emph{{JMLR} Workshop and Conference
  Proceedings}, pages 843--852. JMLR.org, 2015.

\bibitem[Tulyakov et~al.(2017)Tulyakov, Liu, Yang, and
  Kautz]{DBLP:journals/corr/TulyakovLYK17}
Sergey Tulyakov, Ming{-}Yu Liu, Xiaodong Yang, and Jan Kautz.
\newblock Mocogan: Decomposing motion and content for video generation.
\newblock \emph{CoRR}, abs/1707.04993, 2017.

\end{thebibliography}
	
	\newpage
	\appendix
	
	\section{Details of experimental setup}
	\textbf{Moving MNIST}: We downloaded the public available preprocessed dataset from their website \footnote{\url{http://www.cs.toronto.edu/~nitish/unsupervised_video/}}. It consists of sequences of length $20$ where each frame is of size $64 \times 64 \times 1$.
	For the model we learned on the MMNIST dataset, we set the dimension for the static information $\mathbf{f}$ to $64$ and trained it for 60k iterations.  \\
	
	\textbf{Sprites}: To create this dataset, we followed the same procedure as described in \cite{DBLP:conf/icml/LiM18a}. 
	We downloaded the available sheets from the github-repo \footnote{\url{https://github.com/jrconway3/Universal-LPC-spritesheet}} and chose 4 attributes (skin color, shirt, legs and hair-color) to define a unique identity. 
	For each of this attributes we selected 6 different appearances which makes in total $6^4 = 1296$ different combinations of identities. 
	Although, instead of using a single instance of an action sequence, we used the whole sheet which consists of 178 different poses. 
	The size of a single image is $64 \times 64$.
	For the Sprite dataset we increased the dimensions for the latent space  $\mathbf{f}$ to $256$ and trained it for 43k iterations. \\
	
	\textbf{Aff-Wild}: The real world dataset is a preprocessed and normalized version of the Aff-Wild dataset \cite{DBLP:journals/corr/abs-1804-10938}. The dataset consists of 252 sequences, of length between 20 and 450 frames. Instead of using the whole video frame, we cropped the face and resized it to a size of $64 \times 64$.
	
	\section{Details of the encoder for the static latent variable model}
	
	\begin{figure}[h]
		\centering
		\includegraphics[width=0.8\linewidth]{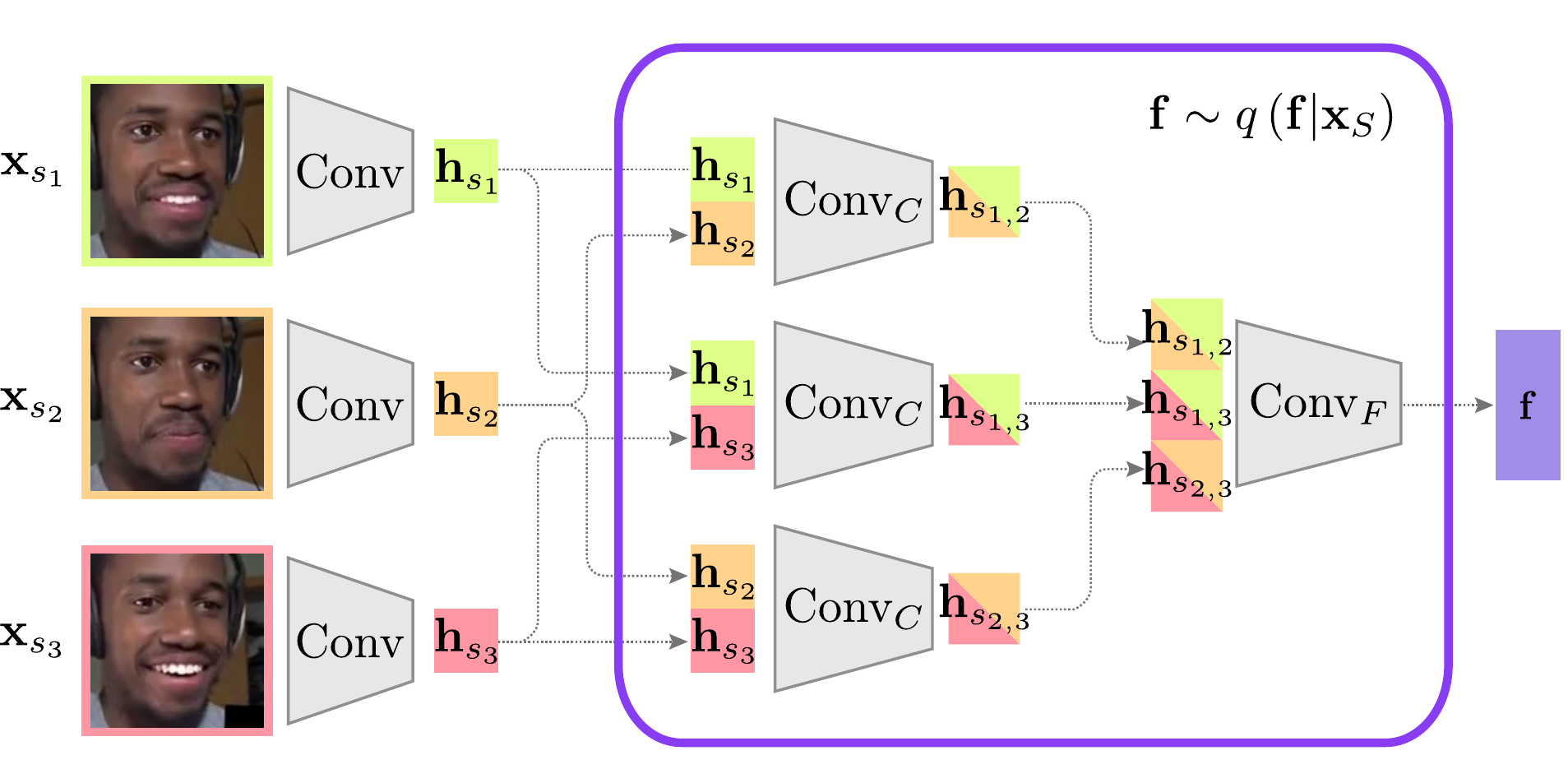}
		\caption{Visualization of the encoder for the static latent variable model}
		\label{fig:encoder_f}
	\end{figure}

	\newpage
	\section{Network Architecture}
	
	\begin{table}[h]
		%Shared Encoder, 
		\begin{tabular}{l}
			\textit{Conv}: $\mathbf{x}_{s_i} \to \mathbf{h}_{s_i}$ \\ \cline{1-1}
			\{ \textit{conv2d:} kernel: 4x4, filters: 256, stride: 2x2, activation: leaky ReLU \}\\
			\{ \textit{conv2d:} kernel: 4x4, filters: 256, stride: 2x2, activation: leaky ReLU \}\\
			\{ \textit{conv2d:} kernel: 4x4, filters: 256, stride: 2x2, activation: leaky ReLU \}\\
			\{ \textit{conv2d:} kernel: 4x4, filters: 256, stride: 2x2, activation: leaky ReLU \}\\
		\end{tabular}

	\vspace{2em}

	\begin{tabular}{l}
		%Merge pairwise frames, 
		$\textit{Conv}_C$: $\operatorname{concat}\left([\mathbf{h}_{s_i}, \mathbf{h}_{s_j}]\right) \to \mathbf{h}_{s_{i,j}}$ \\ \cline{1-1}
		\{ \textit{conv2d:} kernel: 3x3, filters: 512, stride: 1x1, activation: ReLU \}\\
	\end{tabular}

	\vspace{2em}

	\begin{tabular}{l}
		%Merge combined frames, 
		$\textit{Conv}_F$: $\operatorname{concat}\left([\mathbf{h}_{s_{1,2}}, \mathbf{h}_{s_{1,3}}, \mathbf{h}_{s_{2,3}}]\right)$ $\to$ [$\mu_f$, $\sigma^2_f$] \\ \cline{1-1}
		\{ \textit{conv2d:} kernel: 3x3, filters: 512, stride: 1x1, activation: ReLU \}\\
		\{ \textit{dense:} units: 512, activation: ReLU \}\\
		\{ \textit{dense:} units: 1024, activation: None \}
	\end{tabular}

	\vspace{2em}
	%Combine static latent vector with dynamics, 
		\begin{tabular}{l}
			$q_\phi\left(\mathbf{z}_{s_i}\:|\: \dots \right)$: $\operatorname{concat}\left([\mathbf{f}, \mathbf{h}_{s_i}]\right)$ $\to$ [$\mu_z$, $\sigma^2_z$] \\ \cline{1-1}
			\{ \textit{dense:} units: 512, activation: ReLU \}\\
			\{ \textit{dense:} units: 512, activation: ReLU \}\\
			\{ \textit{dense:} units: 4, activation: None \}\\
		\end{tabular}\newline

	\vspace{2em}
		%Learned Prior, 
		\begin{tabular}{l}
			$\textit{p}_\theta\left(\mathbf{z}_{s_i}\:|\:\mathbf{f}\right)$: $\mathbf{f} \to$ [$\mu_z$, $\sigma^2_z$] \\ \cline{1-1}
			\{ \textit{dense:} units: 512, activation: ReLU \}\\
			\{ \textit{dense:} units: 512, activation: ReLU \}\\
			\{ \textit{dense:} units: 4, activation: None \}\\
		\end{tabular}\newline

	\vspace{2em}
		%Deconvolution decoder, 
	\begin{tabular}{l}
		$\textit{Deconv}$: $\operatorname{concat}\left([\mathbf{f}, \mathbf{z_{s_i}}]\right)$ $\to$ $\tilde{\mathbf{x}}_{s_i}$  \\ \cline{1-1}
		\{ \textit{deconv2d:} kernel: 4x4, filters: 256, stride: 2x2, activation: leaky ReLU \}\\
		\{ \textit{deconv2d:} kernel: 4x4, filters: 256, stride: 2x2, activation: leaky ReLU \}\\
		\{ \textit{deconv2d:} kernel: 4x4, filters: 256, stride: 2x2, activation: leaky ReLU \}\\
		\{ \textit{deconv2d:} kernel: 4x4, filters: 3, stride: 2x2, activation: None \}\\
	\end{tabular}
	\end{table}
\end{document}